\documentclass[letterpaper, 10 pt, conference]{ieeeconf}  

\IEEEoverridecommandlockouts                              
\usepackage{graphicx} 
\usepackage{subcaption}
\usepackage[]{cite}
\usepackage{amsmath} 
\usepackage{multirow}
\usepackage[table,xcdraw]{xcolor}

\usepackage[latin1]{inputenc}
\usepackage{tikz}
\usetikzlibrary{shapes,arrows}

\usepackage{algorithm}
\usepackage{algpseudocode}

\title{\LARGE \bf
A Vision-based Scheme for Kinematic Model Construction of Re-configurable Modular Robots}

\author{Kewei Lin, Juan Rojas, and Yisheng Guan$^{1}$.
\thanks{$^{1}$Kewei Lin, Juan Rojas (the corresponding author, Email: juan.rojasn@gdut.edu.cn), and Yisheng Guan are with the Biomimetic and Intelligent Robotics Lab (BIRL), School of Electro-Mechanical Engineering, Guangdong University of Technology, Guangzhou, China, 510006.
        }%
}

\begin{document}

\maketitle
\thispagestyle{empty}
\pagestyle{empty}

\begin{abstract}
Re-configurable modular robotic (RMR) systems are advantageous for their reconfigurability and versatility.
A new modular robot can be built for a specific task by using modules as building blocks.
However, constructing a kinematic model for a newly conceived robot requires significant work.
Due to the finite size of module-types, models of all module-types can be built individually and stored in a database beforehand.
With this \textit{priori} knowledge, the model construction process can be automated by detecting the modules and their corresponding interconnections.
Previous literature proposed theoretical frameworks for constructing kinematic models of modular robots, assuming that such information was known \textit{a priori}.
While well-devised mechanisms and built-in sensors can be employed to detect these parameters automatically, they significantly complicate the module design and thus are expensive.
In this paper, we propose a vision-based method to identify kinematic chains and automatically construct robot models for modular robots.
Each module is affixed with augmented reality (AR) tags that are encoded with unique IDs.
An image of a modular robot is taken and the detected modules are recognized by querying a database that maintains all module information.
The poses of detected modules are used to compute: (i) the connection between modules and (ii) joint angles of joint-modules. Finally, the robot serial-link chain is identified and the kinematic model constructed and visualized.
Our experimental results validate the effectiveness of our approach.
While implementation with only our RMR is shown, our method can be applied to other RMRs where self-identification is not possible.
\end{abstract}
\section{INTRODUCTION}\label{sec:Introduction}
\begin{figure}
	\centering
		\includegraphics[width=8.0cm]{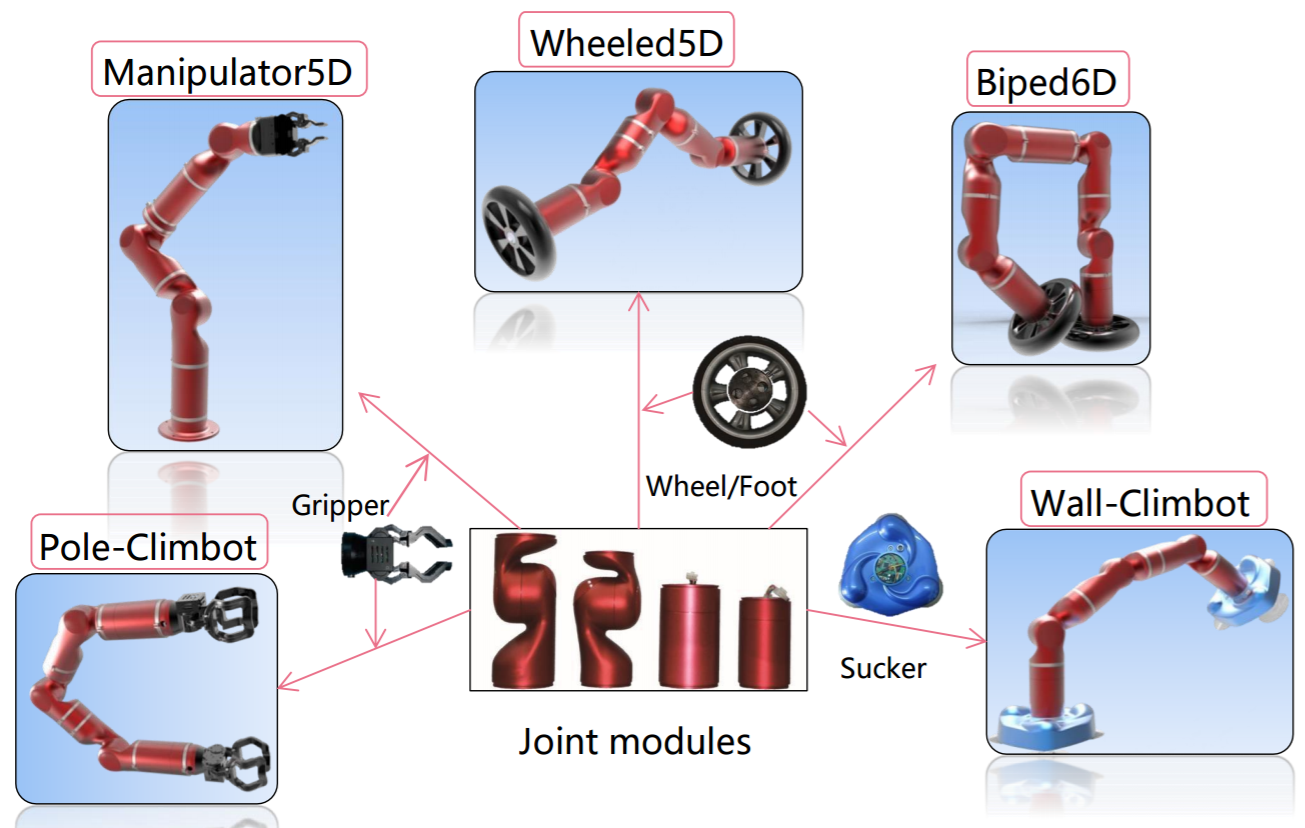}
		\caption{Different kinematic configurations are possible by combining different robot modules in different ways. Also we classify module types in two categories: joint-modules and tool-modules. Joint-modules build the main body of the robot, while tool-modules act as end-effectors.}
		\label{fig:MRS-overview} 
\end{figure}

Modular robotic systems (MRSs) use robotic modules as building blocks to create a variety of kinematic configurations, see Fig. \ref{fig:MRS-overview}. Each module is an independent mechatronic subsystem with a single function. Modules can be of different types. A new kinematic configuration can be specifically selected for a given task and built with modules of different types. This system design principle brings great versatility and flexibility. Throughout this paper, the phrase "kinematic configuration" refers to the robot morphology, whereas the word "configuration" indicates the robot state.

MRSs can be self-reconfigurable robots (SRRs) or not. Unlike SRRs, e.g. Roombot \cite{sprowitz2014roombots:} and M-TRAN \cite{kurokawa2008distributed}, re-configurable modular robots (RMRs) tend to be industry-oriented and are set-up manually by users for specific tasks \cite{Guan2009Development,Chen1999Kernel,Bischoff2010The}. RMRs reconfiguration is conducted by users, whereas SRRs use complex self-docking mechanisms to ensure reliable mechatronic connections between modules and enable self-reconfiguration \cite{Baca2014ModRED}. RMRs are simpler and of lower-cost compared to SRRs. Nevertheless, they share similar challenges in design and control \cite{Yim2007Modular}.

To use a RMR for a specific task, one must (i) set up the physical modular robot and (ii) construct the robot model in the software. Physical connection of modules has been well addressed through the use of quick-lock mechanisms in each module. Modular robots are now set up conveniently with little effort.
The bottle neck of RMRs application lies in constructing kinematic models for newly conceived modular robots, which takes effort and expertise \cite{Chen1999Kernel}.

When considering kinematic configurations of a RMR system, there is a large number of possible combinations given a finite-set of modules and module-types. It's cumbersome for RMR designers to derive all possible kinematic models. Consequently, the user needs to derive a kinematic model every time a new modular robot is conceived. The automation of the kinematic model construction is crucial to further lower the barriers for non-experts.
 
With regard to automatic model generation, Chen \textit{et al.} \cite{Chen1999Kernel} introduced a theoretical framework for automatic RMR model generation. This work assumed that the robot structure was known \textit{a priori} or specified manually.
Rebots, a physics-based simulator for MRS, provides an intuitive way to specify modular robot structures \cite{CollinsReBots}. Given a physical modular robot, a user can observe the module-types and their ordering, and reproduce it in the simulation using drag-and-drop interaction. Despite the intuitive interface, untrained users may mistake the module-types or their ordering.

The goal then is to automate both the identification of the robot structure as well as the kinematic model generation. In doing so, model errors will be minimized and barriers-to-use will be lowered.
For developing plug-and-play RMRs, modules should be able to self-identify: (i) \textit{the order} in which they are connected and (ii) \textit{how} adjacent modules are connected.
There are in fact, self-reconfigurable modular robots that are able to perform self-identification of their kinematic structures. They do so, through built-in sensors that detect the adjacent modules and the connection between modules. M-TRAN \cite{kurokawa2008distributed}, uses IR communication between neighboring modules to learn connection information. The hardware solution (i.e. the use of additional built-in sensors) increases the design complexity of the docking interface, increases costs, and yet the information query is done infrequently. For industrial-oriented RMRs, reconfiguration happens only occasionally (only when new tasks are given), so the inclusion of additional built-in sensors is less preferred.


In this paper we study whether less costly alternatives exist to autonomously identify the robot kinematic configuration and subsequently generate the kinematic model of the system. To this end, we contribute a vision-based kinematic-configuration identification system and an accompanying kinematic model generation scheme for RMRs.
Our work places one or more markers (QR codes, laser engravings, etc) in each module. Each marker has a unique ID.
A camera takes a picture of the modular robot.
Markers are detected by the marker tracking method and their poses are used to estimate the poses of the modules.
By querying the module database, other information of the corresponding modules is also available.
Then with this set of detected modules and their pose estimates, the kinematic structure grows by iteratively (i) finding the connected module for each module and (ii) computing their connection parameters. After identifying the modules and their connections, the robot kinematic chain is constructed.
Moreover, joint-modules' joint angles are also computed to better visualize the generated robot model.
The effectiveness of our approach is demonstrated by the correct model generation of multiple robots with different kinematic configurations or a robot at different configurations.

This paper is organized as follows: In Sec. \ref{sec:framework}, the general identification framework for tree-type RMRs (chain type as a special case) is introduced. 
Sec. \ref{sec:Rep_Modular_Robots} and Sec. \ref{sec:identification} detail the implementation of our method specifically for our RMR system.
In Sec. \ref{sec:Rep_Modular_Robots}, terminology and representation of RMR are introduced. In Sec. \ref{sec:identification} kinematic chain identification and model construction are described.
In Sec. \ref{sec:expriment} experiments are shown. In Sec. \ref{sec:conclusion} discussion of the topic is presented.
\section{GENERAL FRAMEWORK OF KINEMATIC CHAIN IDENTIFCATION}\label{sec:framework}
\begin{figure}
\centering
\includegraphics[width=8.5cm]{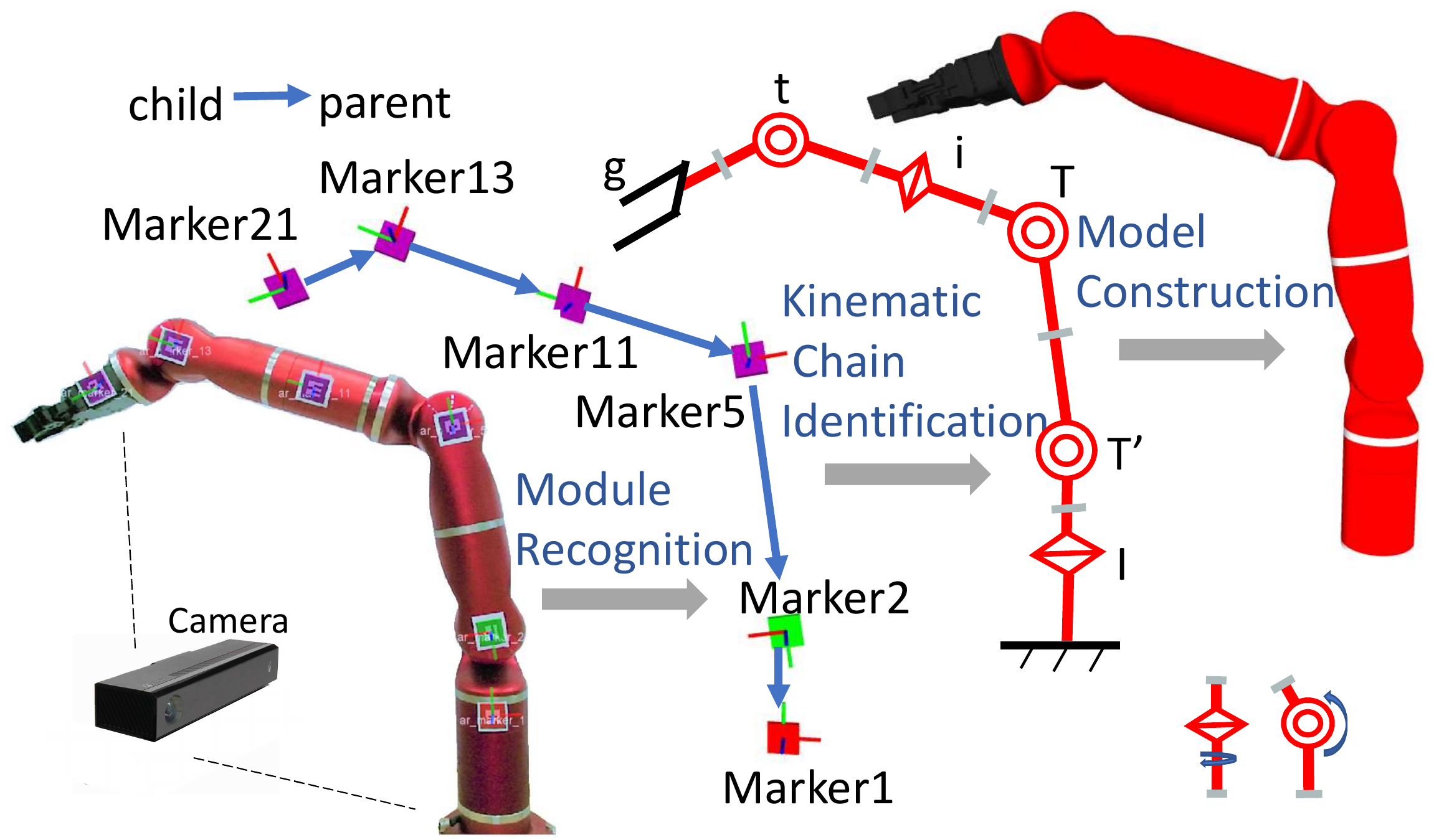}
\caption{Each module is affixed with AR tags. The camera takes an image of the newly built modular robot, a manipulator in this case, for identification of the kinematic structure. Then the robot model is constructed and visualized. }
\label{fig:construction}
\end{figure}

In our method, a camera is used to take images of a newly built modular robot (see Fig. \ref{fig:construction}).
AR-tag-tracking methods (e.g. ar\_track\_alvar \cite{arTrackAlvar}) can be employed to detect marker tags on a modular robot and estimate their poses.

A module database maintains data of all module-types and information of all fabricated modules.
Querying the module database with marker IDs retrieves information of the corresponding modules, including the module-types and communication IDs on the field-bus.

Then the procedure of kinematic structure identification begins to find all parent-child relations among detected modules.
In a pair of  connected modules, the parent module is the module that is closer to the base of the chain, and the child is the remaining one.
A tree-like kinematic structure has one or more modules as its end-effectors, which are called tool-modules, in order to perform useful operations.

\begin{figure}
\centering
\includegraphics[width=7cm]{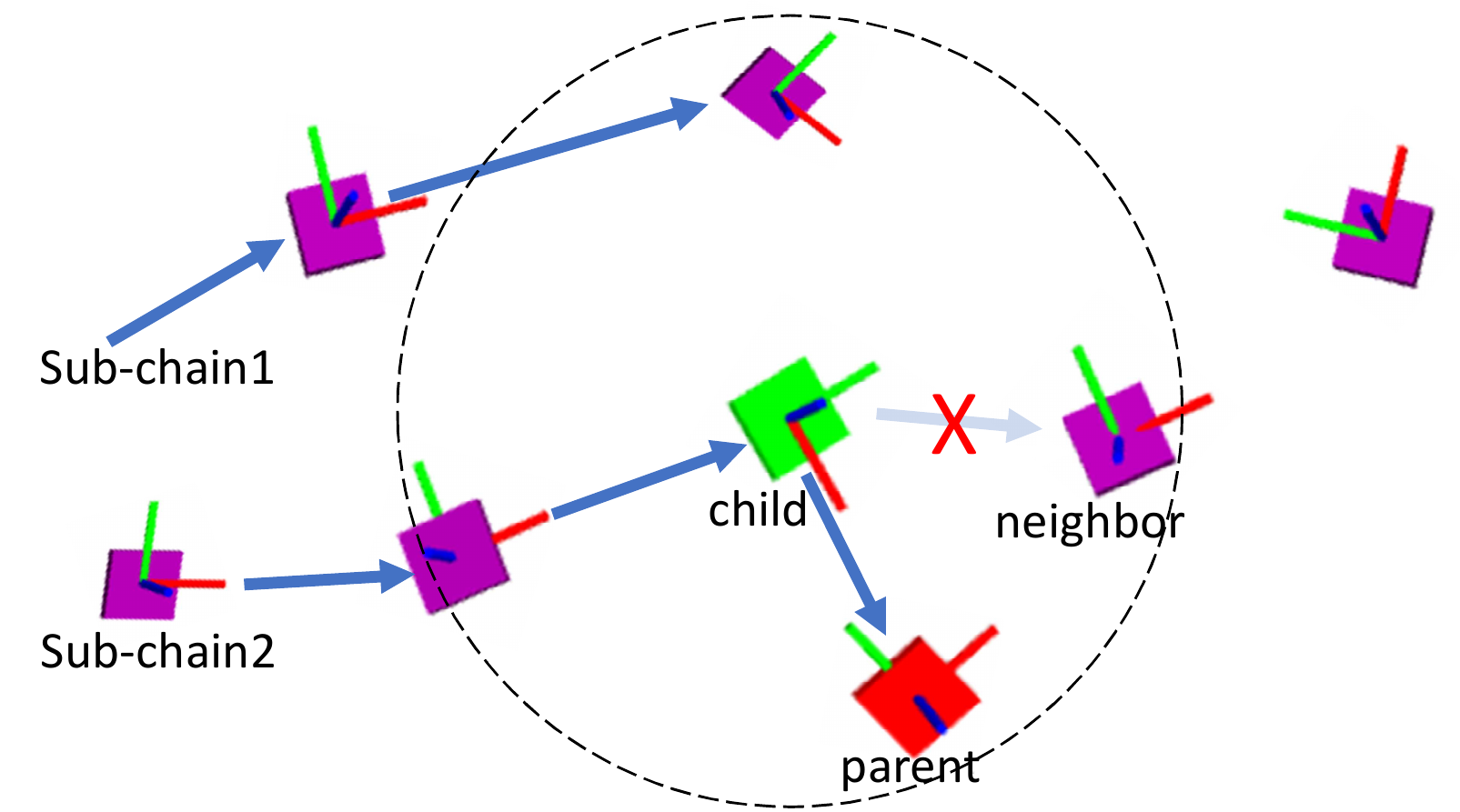}
\caption{Kinematic sub-structures extend themselves by iteratively finding a parent module among detected modules for a child. In the end, all sub-structures form an overall kinematic structure of a modular robot.}
\label{fig:structure}
\end{figure}

An iterative algorithm will identify parent-child relations to build the kinematic chain (see Fig. \ref{fig:structure}).
The algorithm starts by selecting a tool-module as a child and finding its parent. For each succeeding cycle, parent modules become child modules. The chain grows until no further parents are found. This is considered a kinematic branch-chain. In cases, where other branch-chains exist, those would be processed as well resulting in an overall tree-structure for the modular robot.  
For chain-like RMR systems, since there is only one branch chain, the algorithm computes the chain in one pass.

As our approach deals with tree-type structures, the parent-searching method avoids multiple results that appears in a child-searching method, because in a tree-structure a module might have multiple children.

One way to search for a parent module of a child is to exploit the geometric constraints of all module-types, as we will detail in Sec. \ref{sec:identification}.
 
Another alternative is to formulate the problem as an optimization problem. In doing so, a more generalized framework for different RMRs can be established. What's more, with a well-devised cost function, all variables needed to extend the kinematic structure can be found simultaneously, including the parent module, the connection and the state of the parent. The dimension of optimization space does not grow as the degrees of freedom of a modular robot increase, because each time only variables of one module is optimized.

For two modules $m_1$ and $m_2$, relevant definitions are listed in Table \ref{tab:nomenclature}.
\begin{table}[h]
\caption{NOMENCLATURE}
\label{tab:nomenclature}
\begin{center}
\begin{tabular}{cl}
\hline
$M$&a set that contains all modules types\\
$t_{m_1}$&module-type of $m_1$, query  from module database\\
$H_{m_1}$&detected pose of $m_1$ as a homogeneous matrix\\
$O_{m1}xyz$&coordinate frame of $H_{m_1}$\\
$\theta_{m_1}$&module state, e.g. a  joint angle for a joint-module\\
$c_{m_1\ ,{m_2}}$&connection variable between parent $m_1$ and child $m_2$\\
$N_{m_1}$&neighboring modules around $m_1$\\
\hline
\end{tabular}
\end{center}
\end{table}

For most RMR systems, only finite possible connection approaches between two modules are supported: $c_{m_1,m_2}\in \{c_1, c_2, \cdots, c_{n}\}$.

Given a child module, denoted as $c$, the algorithm tries to find out its parent $p$. To do so, first, neighbors around $c$, $N_c$, are found. The distance between a neighbor $n$ and $c$ must not exceed the maximum distance between two connected modules of any module-types and therefore satisfies Eqtn. \ref{eq:neghbors}:
\begin{equation} \label{eq:neghbors}
\left\| \vec{O_{n}O_c}  \right\| \leq max(\left \| \vec{O_{p}O_c} \right \|) + \epsilon_1,\ \forall n\in N_c\ .
\end{equation}
where $\epsilon_1$ is an error threshold.

For each neighbor $n\in N_c$, the transformation from $n$ to $c$ is defined as a homogeneous matrix $T_{c}^n=H_n^{-1}\ H_c$.
If $n$ is the parent of $c$, i.e. $p=n$, a module state $\theta_{n}$ and a connection variable $c_{n,c}$ that transforms the pose of $c$ to that of $n$ can be found.
This transformation is defined as 
\begin{equation} \label{equ:trasformation}
T(c_{n,c}, \theta_c, t_c,\theta_{n}, t_{n})=T_1(\theta_c,t_c)T_2(c_{n,c})T_3(\theta_n,t_n)
\end{equation} where $T_1, T_2$ and $T_3$ denote respectively the transformations relevant to child's state, connection between parent and child, and parent's state.

In other words, if $p=n$, there is a $\theta_n$ and a $c_{n,c}$ that allows $T$ in Eqtn. \ref{equ:trasformation} to theoretically equal $T_c^n$.
A metrics $F$ is devised for measuring the distance between $T$ and $T_{c}^n$:
\begin{equation}\label{equ:metrics}
\begin{split}
F&(T, T_c^n)=\left \|W\circ(T-T_c^n)\right \| _2 \ ,\\
W&=\begin{bmatrix}
w_o &  w_o&  w_o& w_t\\ 
 w_o&  w_o&  w_o& w_t\\ 
 w_o&  w_o&  w_o& w_t\\ 
 0&0&0&0
\end{bmatrix}.
\end{split}
\end{equation}
where $\circ$ is element-wise product and a weighting matrix $W$ that adjusts weights for translational and orientation differences.

The parent-searching procedure now can be expressed in the form of an optimization problem:
\begin{equation} 
\label{eq:optimization}
\begin{aligned}
min:&\ F(T,T^n_c)\\ 
variables:&\ n,\ c_{n,c},\ \theta_n\\
s.t.&\ n\in N_c;\\
&\ c_{n,c}\in \{c_1, c_2, \cdots, c_{n}\};\\
&\ \theta_n\in [min(\theta_{t_c}),\ max(\theta_{t_c})]\ .
\end{aligned}
\end{equation}
If $min(F(T,T^n_c))$ is within a certain threshold, the parent $p=\underset{n\in N_c}{arg\ min}\ F(T, T_c^n)$ is found, as well as the connection variable with its child and the parent state. This parent-searching procedure extends the kinematic chain to next module. And one module by another, the kinematic chain can be completed.
\section{TERMINOLOGY AND REPRESENTATION OF RE-CONFIGURABLE MODULAR ROBOTS}\label{sec:Rep_Modular_Robots}
The following two sections will detail the implementation of our method specifically for our RMR system in \cite{Guan2009Development}.
This section introduces the representation of modules and modular robots in \cite{Guan2009Development}.

\subsection{Re-configurable Modular Robotic System}
Our system and its design methodology are well presented in \cite{Guan2009Development}. Fig. \ref{fig:MRS-overview} shows five robotic systems built with these modules for different applications\cite{Guan2009Development}, \cite{Guan2009A}, \cite{Guan2013A}.

\begin{figure}
\centering
\includegraphics[width=8.0cm]{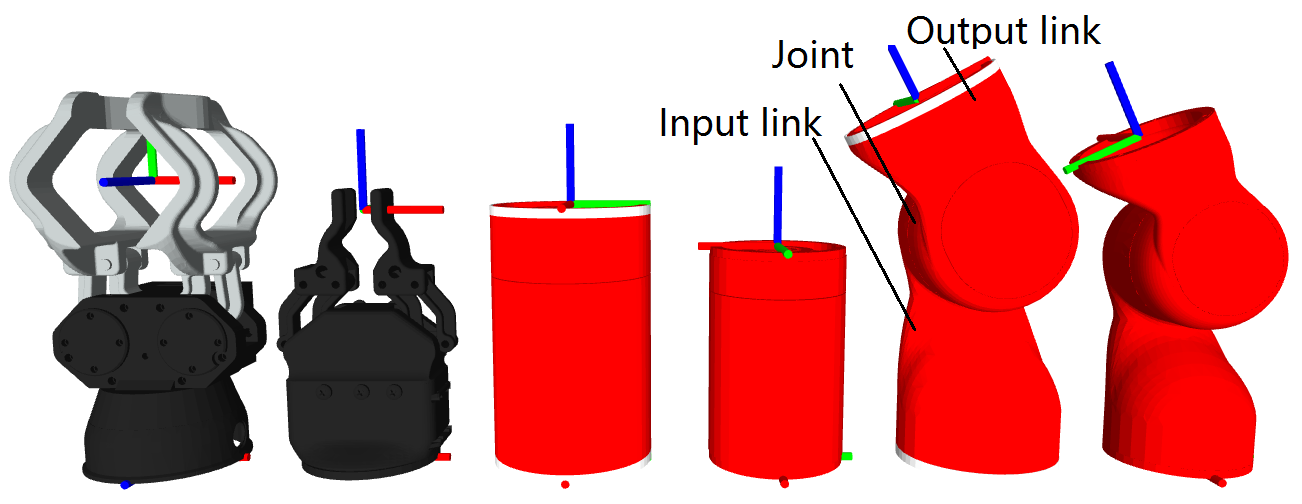}
\caption{From left-to-right, we have two different types of tool-modules that serve as grippers, followed by 4 joint-module types. The first two are simple cylinders that allow for revolute rotations, the last two are composed of two revolute joints and serve as T-connectors. They are denoted as \textit{G}, \textit{g}, \textit{I}, \textit{i}, \textit{T} and \textit{t}.}
\label{modules}
\end{figure}

For joint-modules, upper-case letters \textit{T} and \textit{I} indicate I-typed and T-typed joint-modules whose rotation axes are collinear with and perpendicular to the link axes, respectively. As for tool-modules, we use \textit{G},  \textit{W} and \textit{S} to represent respectively the gripper,  wheel and suction modules. Fig. \ref{modules} shows some of these modules.

Smaller-sized modules are also developed. Using them instead of larger ones near the end of a serial modular robot help ease the torque requirement of its base module. These modules are denoted with lower-case letters such as \textit{t},  \textit{i} and \textit{g}, which indicate the smaller version of \textit{T}, \textit{I} and \textit{G}.

Besides joint-modules and tool-modules, there are accessory modules. Link modules are used to extend the distance between two adjacent joint axes and are denoted as \textit{L} or \textit{l}. Adapter linkage modules, denoted as \textit{A}, are use to connect larger modules with smaller modules.

Most of these modules can be installed in an inverted way in a kinematic chain, i.e. the output link is closer to the base. A single quote after the module-type letter indicates this type of installation, e.g. \textit{T'}.

\subsection{Representation of Basic Modules}
Each module-type is stored in the module database as a Xacro format file that contains kinematic and dynamic parameters, as well as transmission and motor specifications. These Xacro files are named with the corresponding module-type names.

Since the joint-modules can be installed into a kinematic chain in an upside-down way, the terms \textit{input} and \textit{output} are adopted only in the sense of where the driving motors are placed.
To address the problem of inverted installation of modules, Xacro files of inverted modules are also created.

The usage of Xacro files incorporates other useful tools from Robot Operating System (ROS) such as the general inverse kinematics solver TRACK-IK \cite{Beeson2015TRAC}.
\subsection{Representation of Modular robots}
With a database that stores Xacro files of all module-types, we can develop a high-level abstraction of kinematic chains of modular robots by specifying the following information:
\begin{enumerate}
\item constituent modules and their types,
\item the order of each module in the kinematic chain,
\item connectivity between a pair of adjacent modules (connection angle $c$).
\end{enumerate}
Besides, in order to better visualize the robot, we will also estimate the robot state: 4) joint angles of joint-modules.

A text-based method is employed to implement this abstraction for our serial modular robotic system. 
The modular manipulator in Fig. \ref{fig:MRS-overview} is described with a string "\textit{I-T0-T0-A-i0-t180-g90}".
And for the biped tree/truss climbing robot Pole-Climbot in Fig. \ref{fig:MRS-overview}, it is "\textit{G'-I0-T'0-L0-T'90-T180-I'0-G0}". However, it is also appropriate to describe it with "\textit{G'-I0-T'0-T'180-L'(-90)-T0-I'0-G0}" because of its bipedal locomotion behavior.
These strings are organized in the order from the bases of kinematic chains to the ends.
The hyphens are delimiters between adjacent substrings. Each substring contains the module-type and the connection angle $c$ in degrees. The connection angle describes how a module is connected to its parent module. Due to four pin holes on the connection interface of each module, there are only four possible connection angles: $c\in \{-90^{\circ}, 0^{\circ}, 90^{\circ}, 180^{\circ}\}$.

Once the above high-level parameters are specified, a modular robot can be represented with an overall Xacro file, which incorporates macros of the constituent modules according to these parameters.

While straightforward enough for MRS designers, the text-based method still requires users to manually specify the robots structure, which is error-prone for users.
\section{KINEMATIC CHAIN IDENTIFICATION}\label{sec:identification}
To identify the aforementioned high-level parameters, we propose a vision-based method. An open-source project called ar\_track\_alvar \cite{arTrackAlvar} is employed to track AR tags on modules.
\subsection{Placement of Tags}
Every module is affixed with marker tags encoded with unique IDs. 
The markers are placed on the surface of each module such that at least one of them can be captured by the camera while we seek to keep the number of markers to minimum.
Markers on the same rigid body are called a "bundle". The usage of multi-tags bundle allows to estimate the pose of a multi-sided object, even when some of the tags cannot be seen.
One marker in each bundle is designated as the master marker. The poses of the detected markers are transformed into the pose of the master to represent that of the object.

Each I-type module has two bundles, one for the output link and the other for the input link. This allows computing the joint angle $\theta$ using the difference between the poses of two linkages $T^{in}_{out}=H_{in}^{-1}\ H_{out}=Rot(y_{in},\theta)Trans(y_{in},\ dis)$, where $dis$ represents the translational distance between two poses.

For other module-types, only input links are attached with tags in order to reduce the number of tags. For a T-type module, the joint angle can be computed by comparing the pose of its input link with that of the adjacent modules.
 
 A virtual marker, locating at the center of each module with its Y-axis collinear with the axis of input link, is designated as the master marker of the module. For each T-type module, the Z-axis of the virtual master marker is collinear with the joint axis. Using these virtual master markers to represent  poses of modules are convenient for calculating high-level parameters. Now the Y-axis or its reverse of a module is pointing to the parent module, with the only exceptions of inverted T-type modules ${T', t'}$. This allows the use of a geometric method instead of the optimization method for our system.

\begin{figure}
\centering
\includegraphics[width=7cm]{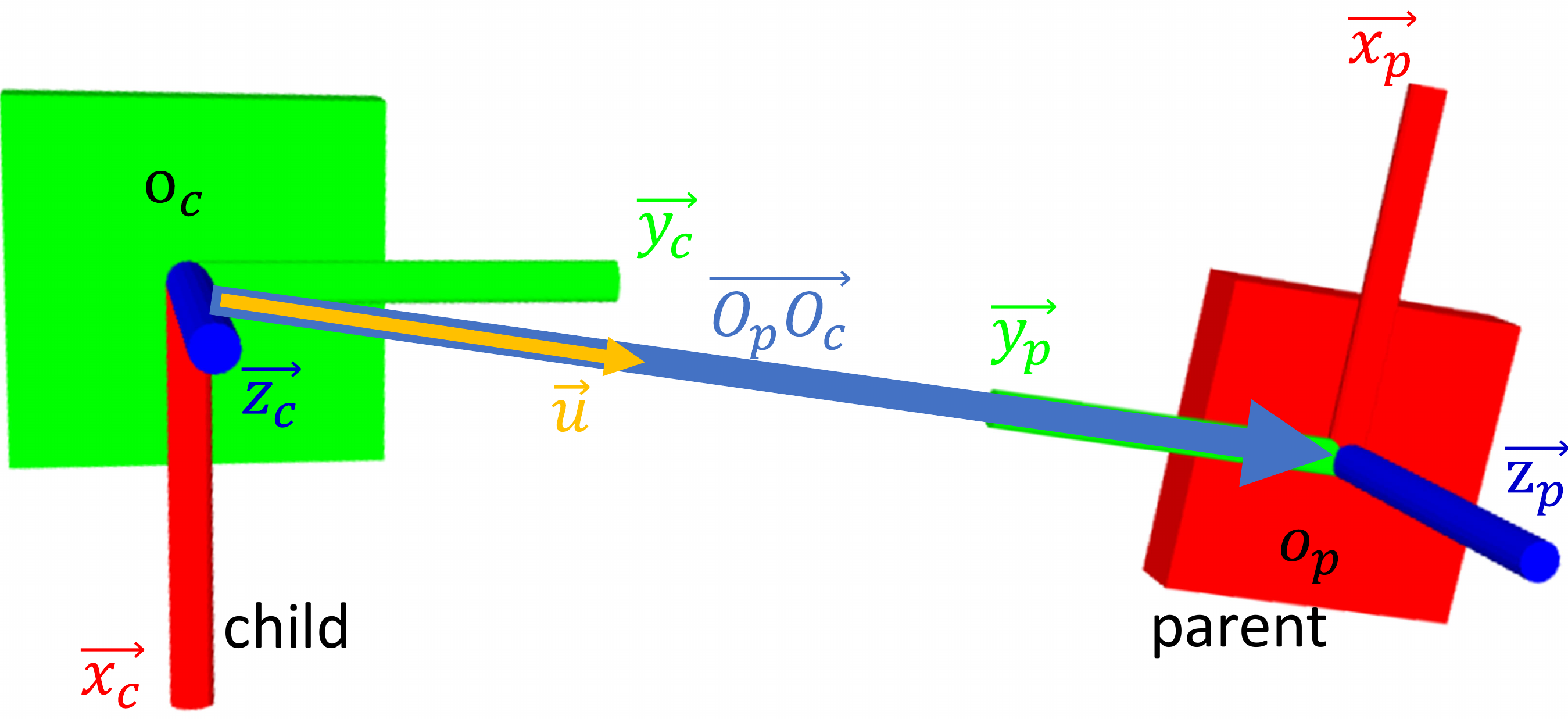}
\caption{A green rectangle represents the master marker of the child module $c$ while the red one represents that of the parent $p$. The frame $O_cxyz$  represents the input link's pose of $c$ while $O_pxyz$ represents that of $p$. $\vec{u}$ is the unit vector of $\vec{O_pO_c}$. }
\label{fig:parent-child}
\end{figure}

\subsection{Module Identification Using Markers}
The ID of each master marker on a module represents the serial number of this module, which is unique and logged in a database that maintains all product information of the fabricated modules. The product information database also manages all the module Xacro files.

Once the markers are detected, they are transformed to the corresponding virtual master markers. By querying the module database, the module-type, ID number on communication bus and other information of each marker are available for further computation.

\subsection{Geometry Constraints}

Our RMR system shows an obvious geometric characteristic: the y-axis of a module's pose usually points to or opposes to its adjacent module, with exceptions when T-type modules get involved.
Therefore, a geometry-based method will be more convenient for our implementation.

The markers tracking and database querying procedures return the poses of constituent modules of a modular robot, which must satisfy a series of geometry constraints. Some useful constraints are deduced as below. 

Here define some symbols for the convenience of discussion. Modules installed in an upright way in the kinematic chain belong to the set $m=\{T, t, I, i, G, g, W, S, L, ,l, A\}$, while the inverted ones belong to $m'=\{ T', t', I', i', G', g', W', S', L', l', A'\}$. A complete set of all module-types is defined as $M=m+m'$.

Consider a pair of connected modules, parent module $p\in M$ and its child module $c\in M$, geometric constraints between them can be organized as a constraint matrix $CST$ in Eqtn. \ref{eq:ConstraintMatrix} according to their module-types. Each element in $CST$,  $cst_{p,c}$ indicates the constraint set that corresponds to the parent and child module-types.

\begin{equation} \label{eq:ConstraintMatrix}
CST=
\begin{bmatrix}
 cst_{T,T} &cst_{T,t} & \cdots   & cst_{T,i'} \\ 
 cst_{t,T}&cst_{t,t}&\cdots &\vdots \\ 
 \vdots &\vdots&\ddots&\vdots \\
 cst_{i',T}& \cdots & \cdots &cst_{i',i'}
\end{bmatrix} .
\end{equation}
As illustrated in Fig. \ref{fig:parent-child}, the frame $O_pxyz$  represents the input link's pose of the parent  module while $O_cxyz$ represents that of the child. $\vec{u}$ is the unit vector of $\vec{O_pO_c}$. 

For practical implementation, only some handy constraints, denoted as $cst$, are used and they are listed as follows: 
\begin{equation}
\begin{split}
cst1:&\ \left\| \vec{O_pO_c}  \right\| \leq max(\left \| \vec{O_pO_c} \right \|_{p,c\in M}) + \epsilon_1\ .\\
cst2:&\ \left|\vec{y_p}\cdot\vec{u}\right|\geq1-\epsilon_2\ .\\
cst3:&\ \left|\vec{y_c}\cdot\vec{u}\right|\geq1-\epsilon_2\ .\\
cst4:&\ \vec{y_p}\cdot\vec{u}\geq 0\ .\\
cst5:&\ \vec{y_p}\cdot\vec{u}<0\ .\\
cst6:&\ \vec{y_c}\cdot\vec{u}\geq 0\ .\\
cst7:&\ \vec{y_c}\cdot\vec{u}>0\ .\\
cst8:&\ p\in m\ .\\
cst9:&\ p\in m'\ .\\
cst{10}:&\ c\in m\ .\\
cst{11}:&\ c\in m'\ .\\
\end{split}
\end{equation}
where error thresholds are defined as: $\epsilon_1\ll max(\left \| \vec{O_pO_c} \right \|_{p,c\in M}), \epsilon_2\ll 1$.
These constraints are summed up in Table \ref{tab:geometryConstraints}  for short. 
 \begin{table}[h] 
\centering
\caption{Geometric constraints of a pair of connected modules.} \label{tab:geometryConstraints}
\begin{tabular}{|c|c|c|c|c|}
\hline
\multicolumn{2}{|c|}{\multirow{2}{*}{$child\setminus parent$}} & \multirow{2}{*}{$p\not\in\{T, t\}$} & \multicolumn{2}{c|}{$p\in\{T, t\}$}          \\ \cline{4-5} 
\multicolumn{2}{|c|}{}                                                      &                             & if $cst6$                   & if $cst7$                   \\ \hline
\multirow{2}{*}{$c\not\in\{T', t'\}$}        & if $cst4$             & $cst1, 2, 3, 8$   & \multirow{2}{*}{$cst1, 10$} & \multirow{2}{*}{$cst1, 11$} \\ \cline{2-3}
                                             & if $cst5$      & $cst1, 2, 3, 9$ &                             &                             \\ \hline
\multirow{2}{*}{$c\in\{T', t'\}$}        & if $c_4$                 & $cst1, 2, 8$         & \multicolumn{2}{c|}{\multirow{2}{*}{$cst1$}}              \\ \cline{2-3}
                                             & if $cst5$      & $cst1, 2, 9$     & \multicolumn{2}{c|}{}                                     \\ \hline
\end{tabular}
\end{table}

By looking up the above table, the parent module can be selected among the neighbors.
Note that though Table \ref{tab:geometryConstraints} is an incomplete table for $CST$, e.g. $\{cst_1, cst_2, cst_3, cst_8\}\in cst_{t,g'}|_{c_4}$, it's useful for finding parent-child relations in a kinematic chain.

Once the parent module is found, the relative pose $T^p_c$ is exploited for calculating the connection angle $c_{p,c}$ and joint angle of parent $\theta_p$. For connection angle $c_{p,c}\in \{-90^{\circ}, 0^{\circ}, 90^{\circ}, 180^{\circ}\}$:
 \begin{equation}
 \begin{split}
c_{p,c}&=discretize(\hat{c}_{p,c}) \label{equ:discretize}\ ;\\
\hat{c}_{p,c}&=
\begin{cases}
 			\arccos(\vec{z_p}\cdot\vec{z_c})&,\text{if } \vec{z_p}\times \vec{z_c}\cdot \vec{u}\geq 0 \\ 
			-\arccos(\vec{z_p}\cdot\vec{z_c})&,\text{if others} 
\end{cases}
 \end{split}
\end{equation}
\subsection{Kinematic Chain Construction}
\begin{figure}
\tikzstyle{decision} = [diamond, draw, fill=red!20, node distance=1.5cm, text width=4em, text centered, node distance=2.5cm, inner sep=0pt]
\tikzstyle{block} = [rectangle, draw, fill=blue!20, node distance=1.4cm, text width=7em, text centered, rounded corners, minimum height=2em]
\tikzstyle{line} = [draw, -latex']
\tikzstyle{cloud} = [draw, ellipse,fill=yellow!20, node distance=3cm, minimum height=3em]
    
\begin{tikzpicture}[node distance = 1.4cm, auto]
    \node [block] (init) {marker detection};
    \node [cloud, right of=init] (db) {database};
    \node [block, below of=init, text width=12em] (query) {query database on a $marker$};
    \node [decision, below of=query, node distance=2cm] (isMarkerEnd) {marker on end-effector?};
    \node [block, left of=isMarkerEnd, node distance=3cm, text width=4.5em] (next) {next $marker$ in $markers$};
    \node [block, below of=isMarkerEnd, node distance=2.2cm, text width=12em] (findChain) {kinematic chain construction};
    \node [block, below of=findChain, text width=12em] (generation) {robot description generation};
    \node [block, below of=generation, text width=12em] (joint) {calculate joint angles};
    \node [block, below of=joint, text width=12em] (visualize) {model visualization};

    \path[line](init)--node {$markers$}(query);
    \path [line,dashed] (db) |- (query);
     \path[line](query)--(isMarkerEnd);
    \path [line] (isMarkerEnd) -- node [near start] {no} (next);
    \path [line] (next) |- (query);
    \path [line,dashed] (db) |- (generation);
     \path [line,dashed] (db) |- (findChain);
    \path [line] (isMarkerEnd) -- node {yes}(findChain);
    \path [line] (findChain) --  node{$chainList$}(generation);
    \path [line] (generation) --(joint);
    \path [line] (joint) -- node{joint angles}(visualize);
\end{tikzpicture}
\caption{Flowchart of the modular robot identification.}\label{fig:flowchart}
\end{figure}
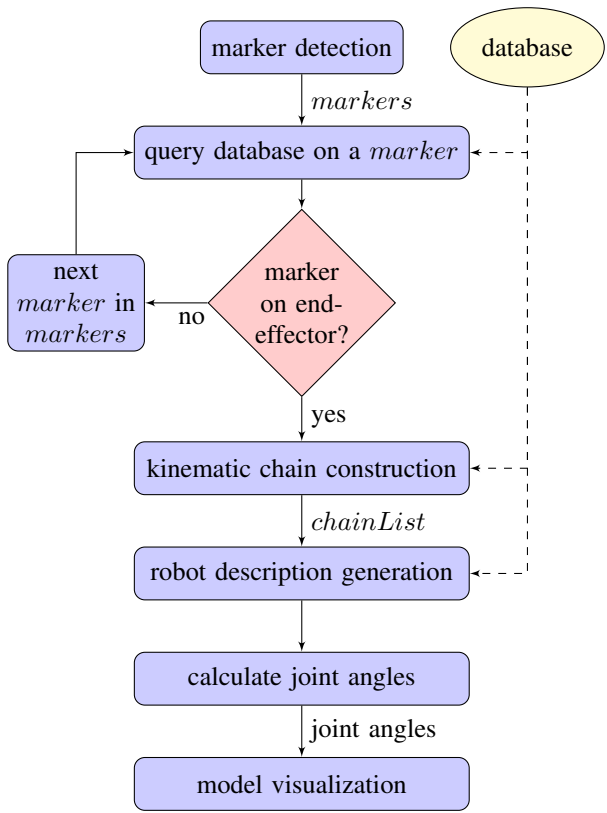

After setting up the physical modular robot, the users launch the robot identification program to start procedures illustrated in Fig. \ref{fig:flowchart}.

By querying the module database, the algorithm examines all detected markers for their validity and, if valid, retrieves their module information. False positives are eliminated while valid markers remain and are exploited to build the kinematic chain. 

Tool-modules are used as the end-effectors. The procedure of kinematic chain identification starts from the first detected tool-modules and grows along the chain to the base module. In the biped climbing robot, two tool-modules are used, either of which can be regarded as the end of the kinematic chain.

Once the end module is found, its pose is used for finding its parent module and calculating the connection parameter according to the geometric constraints in Table \ref{tab:geometryConstraints}. This process continues iteratively until all identified modules find their place in the kinematic chain. A recursive algorithm, see Line \ref{li:recursion} in Alg. \ref{alg:ChainConstruction}, is designed for the process. Note that $chainList$ in Alg. \ref{alg:ChainConstruction} returns the kinematic chain in the order from base to end.

\begin{algorithm}
  \caption{Kinematic chain construction}
  \label{alg:ChainConstruction}
  \begin{algorithmic}[1]
    \State Initialize $chainList$
    \State $childMarker \leftarrow markerOnEndEffector$
    \Function{BuildChain}{$childMarker$, $markers$} 
    \If{markers is empty}
    	\State \textbf{return}
    \EndIf
    \ForAll{$marker \in markers$}
    	\If{$marker$ is parent of $childMarker$}
          \State Find connect angle $c_{p,c}$, install direction $d$.
          \State Remove $marker$ from $markers$.
          \State \textbf{break for loop}
        \EndIf
    \EndFor
    \State BuildChain($marker$, $markers$) \Comment{recursion} \label{li:recursion}
    \State Append [$marker$, $c_{p,c}$, $d$] to $chainList$.
    \EndFunction
  \end{algorithmic}
\end{algorithm}

After identifying the kinematic chain, the joint angles are computed for robot model visualization, which allows the users to intuitively confirm the result.
Robot specification files, as well as files for control, are also generated.
The generated files are essential for our ROS-based control software, which aims to provide a programming framework for modular robots.
Our work on robot model identification paves the path towards the goal of kinematic-configuration-agnostic programming for modular robots.
\section{Experiments}\label{sec:expriment}
\begin{figure}
\centering
\includegraphics[width=8cm]{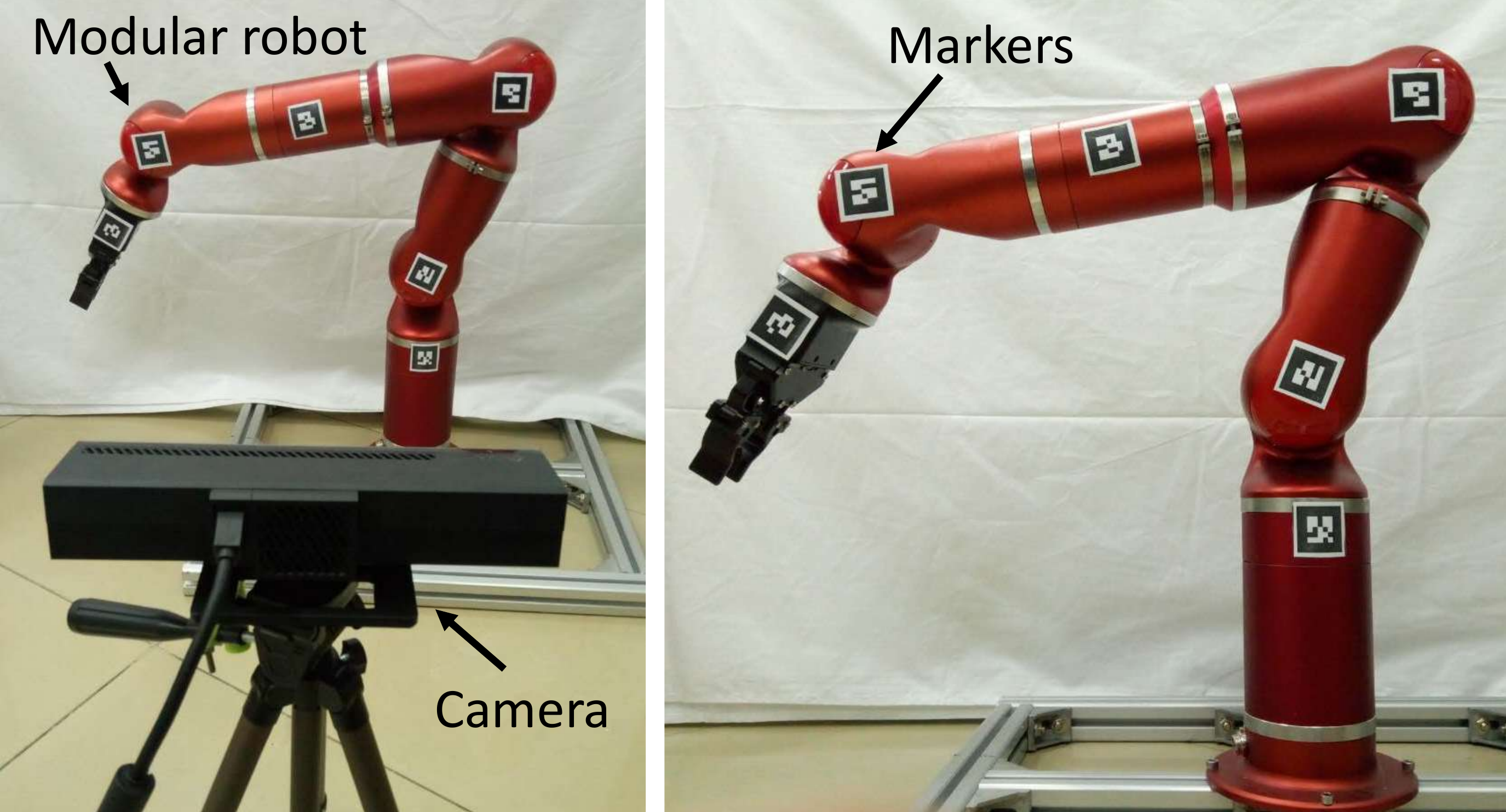}
\caption{Experimental set-up for modular robot identification. Only a camera is needed in our method.}
\label{fig:setup}
\end{figure}
 In our method, only a camera is needed. To be specific, a Kinect is used in Fig. \ref{fig:setup}, but any common camera can be used. Currently AR tags are used, however we envision our modules being marked through laser engravings during fabrication.

\begin{figure}
\begin{subfigure}{.5\textwidth}
  \centering
  \includegraphics[width=7.8cm]{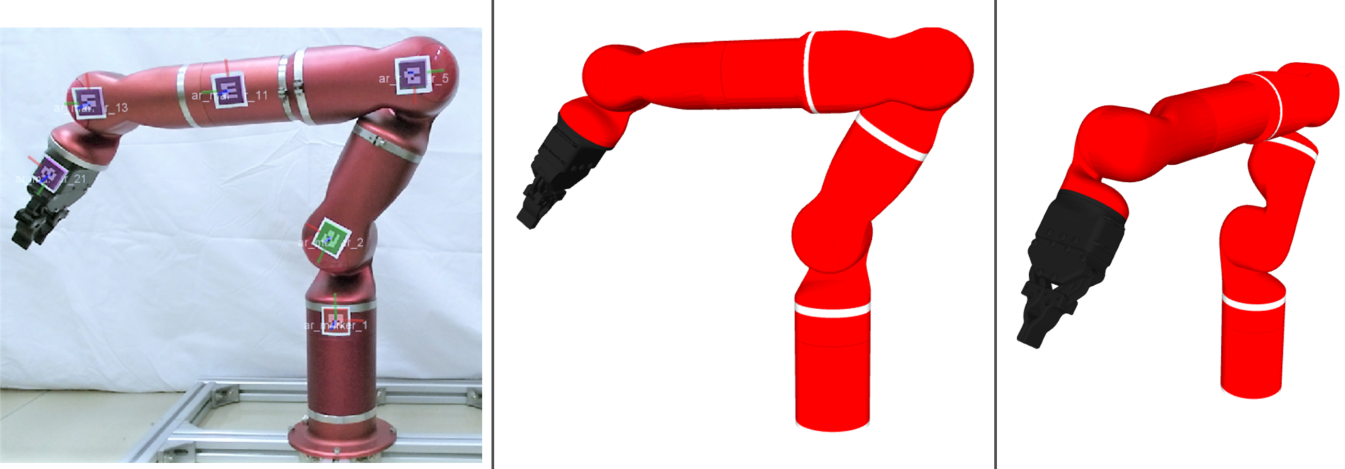}
  \caption{Modular manipulator.}
  \label{fig:experimentA}
\end{subfigure}\\
\begin{subfigure}{.5\textwidth}
  \centering
  \includegraphics[width=7.8cm]{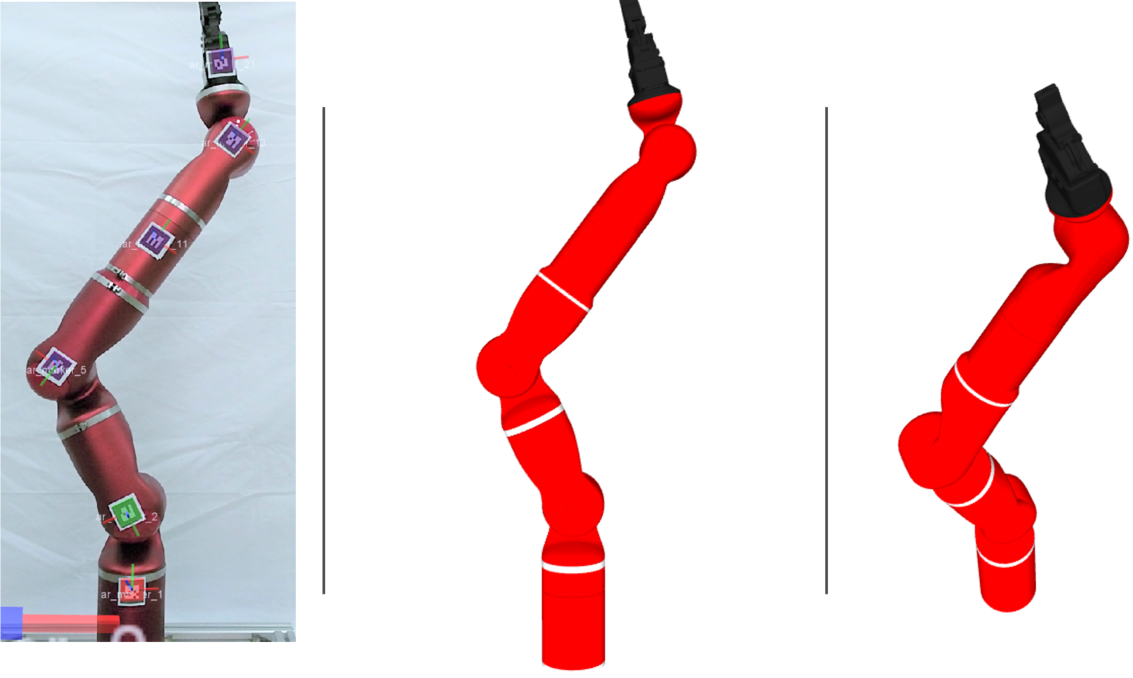}
  \caption{Modular manipulator at a different configurations.}
  \label{fig:experimentB}
\end{subfigure}\\
\begin{subfigure}{.5\textwidth}
  \centering
  \includegraphics[width=7.8cm]{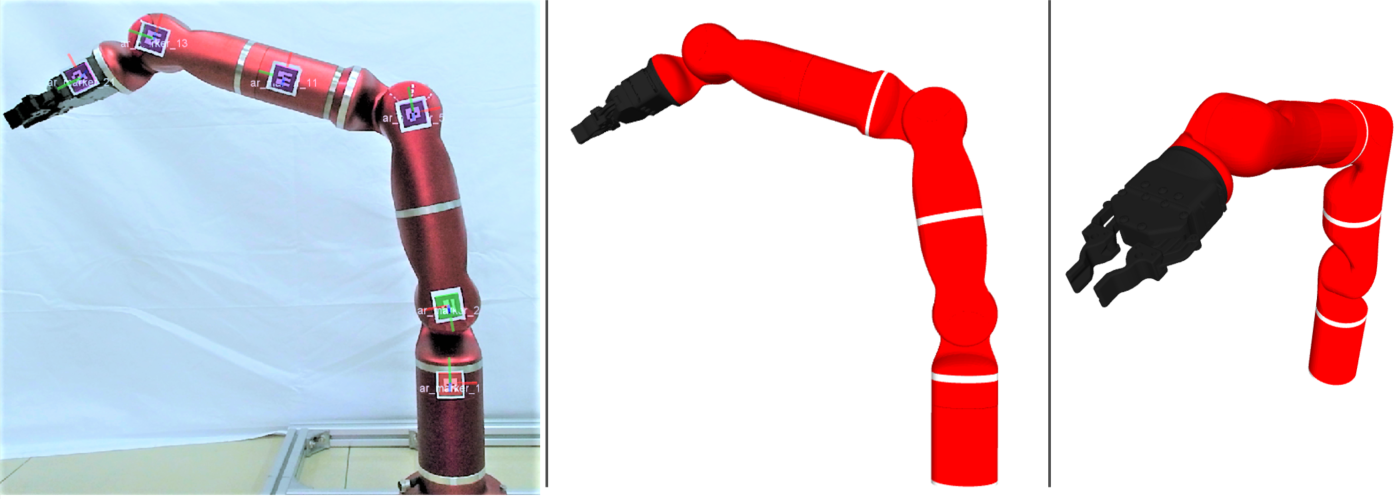}
  \caption{Modular manipulator of different kinematic configuration.}
  \label{fig:experimentC}
\end{subfigure}\\
\begin{subfigure}{.5\textwidth}
  \centering
  \includegraphics[width=7.8cm]{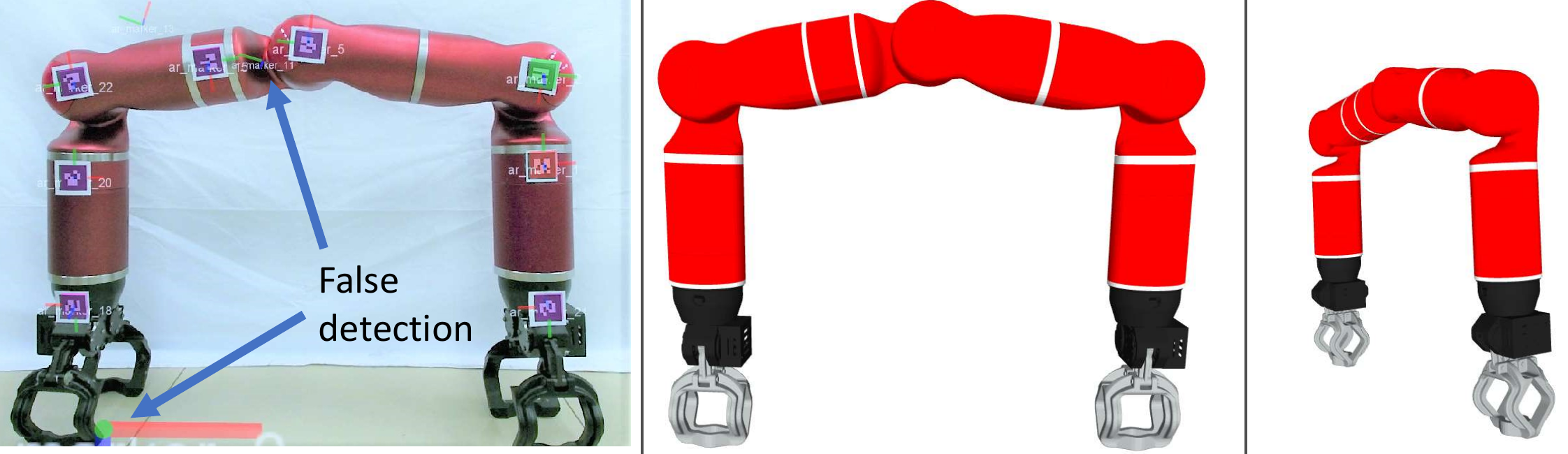}
  \caption{Biped climbing modular robot.}
  \label{fig:experimentD}
\end{subfigure}
\caption{Three modular robots were tested. The first two experiments used the same robot but at different configurations. Despite the existence of false marker detection, the algorithm could identify the kinematic chains and compute the joint angles for visualization.}
\label{fig:experiments}
\end{figure}

Four tests were carried out, as illustrated in Fig. \ref{fig:experiments}.
Our algorithm successfully identified the kinematic chains and computed the joint angles.
In each sub-figure, the left part shows an image of a modular robot, taken by the camera, while the middle and the right columns illustrate the resulting model, which are constructed on-the-fly.

The first two experiments, shown in Fig. \ref{fig:experimentA} and Fig. \ref{fig:experimentB}, used the same modular manipulator, but at two different configurations.
They both shared a same kinematic chain: \textit{I-T'0-T'0-A0-t0-i0-g0}.
The third experiment identified another modular manipulator with a kinematic chain \textit{I-T'0-T0-A0-i0-t0-g0}.
The robot in Fig. \ref{fig:experimentD} is not a manipulator, but a bipedal climbing robot: \textit{G'-I0-T'0-L0-T'0-0T-0I'0-G0}, showing that the proposed method can be applied to bi-directional kinematic chains such biped robots.

Though kinematic chains were correctly identified, joint angle estimations were not precise due to the errors of tag placement on the module and the detection errors.
Some false detections appeared in Fig. \ref{fig:experimentB} and Fig. \ref{fig:experimentD}.
They were eliminated as they failed to match the information stored in the module database, or couldn't find any parent or child.
\section{DISCUSSION AND CONCLUSION}\label{sec:conclusion}
In this paper we showed that kinematic-configuration identification and automatic kinematic model generation can be done on the fly by using a vision-based scheme and a geometric mathematical model of serial-chain systems. 

Reconfigurability brings great advantages and also challenges, comparing to robots with fix-morphology.
The software of RMR systems is supposed to support features corresponding to the reconfigurability.
The rapid development of robot middlewares provides useful tools for implementation of software for RMRs.
Xacro files, widely used in ROS, can be used to represent different types of modules as XML macros.
Generic kinematic solvers such as TRACK-IK and IKFast, with properly specified kinematic structures, provide kinematic solutions for robots.
These can greatly simplify the painstaking construction process of kinematic models, assuming that the kinematic chain is known or manually specified.

To identify the kinematic structures of modular robots allows full automation of the model construction process.
We proposed a vision-based scheme for kinematic configuration identification and model construction of RMR in this paper.
Each module is affixed with AR tags, which can be laser-engraved on the module shell during fabrication.
A camera is used for tracking these markers, whose poses and IDs are used for computing the kinematic chain and joint angles.
Finally, the constructed models are visualized for validation.
All the above procedures are done on-the-fly.
Comparing to solutions that use built-in sensors or other mechatronic device in each module, our method is of lower-cost and can be applied to other modular robotic systems in which self-identification is not possible. 

In the further, in order to thoroughly address the occlusion problem, a hand-held camera can be employed to scan a robot while a multi-view geometry algorithm keeps tracking all marker poses.
What's more, our ultimate goal is a system that allows kinematic-configuration-agnostic programming for RMR systems, i.e. programming is not specifically for a particular modular robot but for all possible ones.


\addtolength{\textheight}{-12cm}   





\section{Acknowledgments} \label{sec:Acknowledgments}
This work is supported by the grants: "Major Project of the Guangdong Province Department for Science and Technology (2014B090919002), (2016B0911006)."
\bibliographystyle{IEEEtran}
\bibliography{IEEEabrv,Xbib} 
\end{document}